\documentclass[11pt,letterpaper]{article}
\usepackage{naaclhlt2015}
\usepackage{times}
\usepackage{latexsym}
\usepackage{hyperref}
\usepackage{url}

\title{Interactive Knowledge Base Population}

\author{
{\bf Travis Wolfe}\ \ \ \ {\bf Mark Dredze}\ \ \ \ {\bf James Mayfield}\ \ \ \ {\bf Paul McNamee}\\
{\bf Craig Harman}\ \ \ \ {\bf Tim Finin}$^{\dagger}$\ \ \ \ {\bf Benjamin Van Durme}\\
Human Language Technology Center of Excellence \\
Johns Hopkins University\\
$\dagger$University of Maryland, Baltimore County
}

\newcommand{\kbpvis}[0]{Quicklime}  
\newcommand{\parma}[0]{Parma}
\newcommand{\slinky}[0]{Slinky}

\renewcommand{\paragraph}[1]{\textbf{#1} }
\date{}

\begin{document}

\maketitle

\begin{abstract}
\vspace{-.3cm}
Most work on building knowledge bases has focused on collecting entities and
facts from as large a collection of documents as possible.  We argue for and describe a new
paradigm where the focus is on a high-recall extraction over a small collection
of documents under the supervision of a human expert, that we call Interactive
Knowledge Base Population (IKBP).  
\end{abstract}

\section{Introduction}

Work on knowledge base population in its various forms (e.g., TREC KBA and TAC
KBP), which we will describe using the umbrella acronym KBP, has primarily
focused on large text collections.  The knowledge base, in the ideal sense, is
the repository for all information captured from this collection of text.
Most KBP work has focused on large, heterogeneous text collections, like Wikipedia or the Internet,
containing a variety of topics, each of which
has its own set of relevant entities, types of facts, writing styles, etc.
This variability and diversity is one of the things that makes entity linking,
relation identification, and other tasks difficult in the large.

In contrast, we are concerned with smaller, more topically focused collections of restricted heterogeneity.
This restricted paradigm changes the KBP problem
significantly, e.g. a system is not
able to rely on data redundancy, as most facts will only be stated once.
This means that a system must address more vague or ambiguous ways of
expressing a proposition than would be considered by a system designed for the
large-text version of the problem.
Yet this setting also offers new opportunities for KBP systems.
In the small text collection paradigm, asking a user for a fairly
complete annotation of the entities and proposition expressed in text is
feasible, and system output can be vetted by an expert. 
We term this paradigm interactive KBP (IKBP). 

The IKBP paradigm can apply to a variety of user types. Roughly
speaking, they are any information analyst whose workflow involves
carefully reading textual information and producing reports (e.g., a
dedicated hobbyist creating a Wikipedia page, a journalist writing a
newspaper article, or a financial analyst generating a quarterly
summary report).  These users can rely on a {\em pocket KB} -- a small
domain or user-specific knowledge base used as a tool for
understanding new sources and producing reports.
Pocket KBs accurately capture the
scope of ambiguity represented in a particular topic of documents (as opposed to a
global KB which will contain many irrelevant entities and facts that introduce
noise into inferences).  This concept of a database that matches a particular
topic or domain opens the door to more speculative inferences and
domain-specific learning that is more computationally 
efficient than would be possible for general KBs.

We describe the IKBP setting, summarize our work in building 
components of an IKBP pipeline, and demonstrate how 
the methods developed for KBP in the general sense can be 
extended to IKBP.


\section{Interactive Knowledge Base Population}

\paragraph{Ideal Users}
Professional analysts across a wide variety of disciplines are tasked with reading and synthesizing
articles on a regular basis.  Examples include financial analysts who need to
keep track of a set of companies and entities relevant to a portfolio,
scientific researchers who must read new articles related to their research,
and public relations staff who track stories relating to their clients. Unlike the average individual,
analysts have a good deal of domain knowledge on their topic, and are willing and able to invest
time and effort to extending their knowledge and developing resources useful for their domain of
interest.
Much of their time is spent reading source articles, identifying
entities of interest, understanding propositions about them, and recalling
these facts while writing a report to synthesize their findings.  These analysts
are experts because of the inferences they can draw from evidence, but their
efficiency is limited by the need to decompress (i.e., read, comprehend, 
recall) information contained in natural language.

Given their investment in the domain, analysts can derive long term benefit from providing minimal annotations
while reading, storing them as structured data to be used
later when a report is synthesized.
Annotations as simple as a bag of entity mentions (coreference) and facts
(relation extractions) offer a powerful index and summary of source documents.
Additionally, this structured data supported by textual mentions offers the
analyst a rich way to cite claims made in a report.
This information captured by the analyst constitutes a lightweight ad-hoc
KB.


\paragraph{Interactive Workflow}
An IKBP system utilizes provided analyst annotations by interesting itself into an analyst's
document-centric\footnote{Here we describe the reports as text documents, but
this need not be the case. IKBP construed broadly works on any source that may
be used to ground out a statement made in a report, such as a spreadsheet,
output of machine translation or speech recognition, or even richer media like
images or video.} workflow, bootstrapping a light-weight knowledge base as fast
as possible.
IKBP is not simply active learning for KBP: while having a human in the loop
providing feedback to the system will indeed allow for updating and improving
models, the users in this case are not primarily concerned with assisting in
building a better system.  Rather, they are consumers of the output of the system
(facts discovered from content) that provide feedback to the system as a
\emph{by-product} of their professional desire to filter and correct output as
a part of their writing process.

To meet the goal of being minimally obstructive to the user, the primary
interface for IKBP is a document viewer. 
During reading, this interface enables a user to make annotations about
entities and relations that are important.  During report writing the interface
should allow for linking statement back to supporting evidence.  In some cases
this may be possible to do fully automatically (e.g. linking a statement about
an entity's birth-date or employer), but in other cases the link must be
explicitly added by the user for the purpose of structured citation (e.g. a
statement that an entity is a `supporter' of a group might be tied back
to mentions of events between that entity and group).
an entity's birth-date or employer),
while in more complex cases the user may opt to make an explicit citation
(e.g. that an entity is a `supporter' of a group might be tied back
to mentions of events between that entity and group).

IKBP should be seen as a natural complement to efforts such
as topic detection and tracking (TDT) \cite{TDT} or knowledge base acceleration (KBA),
which provide tools that triage high-volume content streams to a smaller,
personalized collection that the user can then analyze in depth.


\paragraph{Pocket KBs}
Pocket KBs are task-specific knowledge bases typically associated with a single
user or group of users, such as a group of co-authors working on
a research paper.
First, there are some properties of KBs that will vary from topic to topic and
are poorly modeled by the classic notion of a global KB.  For
example, while
an entity's popularity is an informative prior for the
referent of an unknown mention \cite{Ji2011}, popularity is defined as an
empirical distribution, which will vary greatly across topics.  Any global
distribution over entities will be greatly biased in some domains, and serve as
a poor prior.  Pocket KBs can overcome this bias because they
focus on a coherent topic or domain.  While methods like hierarchical Bayesian
models offer topical specialization and allow for global backoff, they do
not share a pocket KB's benefits of sparsity and compactness.

Another practical issue addressed by pocket KBs is that of ownership and
permissions.  Since they are allocated to a particular user or small group,
curatorial choices do not affect other users (or if they
do it, is clear that the owner of the pocket KB comes first).  KBs that store
global information that affects many users, such as in Wikipedia and Freebase, must use gate-keepers to
moderate changes.  Pocket KBs
are an efficient solution for cases where there is a limited amount of
overlap among users' topics.

Pocket KBs also elegantly handle the issues of scope of ambiguity.
Stoyanov et al.~\cite{CALE} argue for the notion of a context in which a reader can
be expected to perform entity linking, based on Grice's principle of
cooperative communication \cite{Grice1975}.  Their claim is that most entity linking
choices are easy in the correct context. They explored ways to derive the
correct context from a global KB.  While this is a general way to
instantiate contexts to reason about and disambiguate entities, we argue that a pocket KB that
is constructed to only ever include entities that are relevant to a particular
topic is a more efficient way to construct a context.

Lastly pocket KBs sidestep engineering issues around supporting a large global
KB.  The tools needed to do AI research on large KBs like Freebase range from
cumbersome to lacking.  As a result, many state-of-the-art systems are designed
as long pipelines that operate only in batch mode, often requiring days to run
a single experiment \cite{KelvinSlides}.  Working with pocket KBs requires much
less engineering time and allow (non-systems) researchers to run experiments more
quickly.


\section{Relation to Existing Work}
\label{sec:related}



  Since 2009,
the Text Analysis Conference Knowledge Base Population (TAC KBP)
workshops have run competitions on entity linking (determine the
referent of a named entity mention), slot filling (given an entity and
a relation, find string-valued items that complete the relation), and
Cold Start (combine the two previous tasks with no provided KB).  Cold
Start is most similar to IKBP as they both begin with an
empty KB. They differ in that TAC focuses on large
bodies of text and expects offline processing rather than
under the supervision of a user.  Nonetheless, an online optimize Cold Start system
could be the heart of an IKBP system.

Another parallel line of work is the Text Retrieval Conference's workshop on
Knowledge Base Acceleration (TREC KBA).  The Vital Filtering task evaluates how
well a system can link documents in a very high volume document stream (a
billion documents) to entities in a populated knowledge base.  Once documents
are linked to entities, the Streaming Slot Filling task takes these mentions
and updates KB entries with the new information in the linked document.  This
task is relevant to IKBP as it solves the IR task of
finding documents for an analyst to read.

A third line of relevant work is Topic Detection and Tracking (TDT) \cite{TDT}.
Its goal was to find series of articles that constituted a coherent ``news
story'' or topic.  There is significant overlap between this and IKBP,
including cross-document coreference resolution and event detection and
linking.
Like TDT, there is a lot of work in first story detection in social media, such
as Osborne et al.~\cite{Osborne2014}  These systems, while tuned to social media, are also
well suited to linking events as a part of IKBP.

Finally, many systems have addressed tasks similar to IKBP,
such as interactive IE \cite{Culotta2006}, post-hoc diagnostics
of IE errors \cite{Das2010}, and robust cross-document entity
coreference \cite{Minton2011}.
There has been a variety of work addressing the challenges of
knowledge base construction and relation extraction at web-scale
\cite{carlson2010toward,Kasneci,nakashole2011scalable,zhu2009statsnowball},
and DeepDive \cite{niu2012deepdive} in particular emphasizes provenance information and user feedback.
Most significantly, Budlong et al.~\cite{Budlong2013}
built a system to do IKBP for intelligence analysts based on
IE tools such as SERIF \cite{SERIF}; including an interface for
analysts to view and correct annotations. It differed from our IKBP
notion in its focus on entity-centric link analysis and in not stressing
user citation and report writing or exploiting pocket KBs as
valuable aspects of the task.

\section{Our Work on IKBP}

\paragraph{Visualization}
\kbpvis\ is a tool designed to show the user documents that connect to the
knowledge base that we've built so far.  The tool presents content similar 
to what you might see in an online news site, and highlights 
all of the entities and relation triggers produced from within-document coreference resolution and ACE-style
relation extraction systems, with links to KB entries.  These
annotations help the reader skim the document to get an entity or
relation-centric summary or find the source of a particular
fact that the system has asserted. 

\paragraph{Kelvin}
We have adapted Kelvin \cite{McNamee2012,McNamee2013,mayfield-14-aaai}, the
top-ranked system for the TAC Cold Start KBP task in 2012 and 2013, for
building pocket KBs.  Kelvin uses the BBN tools SERIF and FACETS to perform
document level analysis, including detection of named entities, within-document
coreference analysis, and detection of entity relations.  SERIF is an ACE
system, which is more-or-less mappable onto the TAC-KBP schema. FACETS is a
maximum entropy tagger that extracts attributes from personal noun phrases.
Additional within-document coreference analysis is done using data from the
Stanford NLP coref system and by manually developed rules.

\paragraph{Kripke} Cross-document entity coreference clustering is done by Kripke,
which looks for high levels of string matching
(through a number of name-matching metrics) and for contextual
matching (using named entities that are common across source
documents).  Co-occurring named entity mentions and name matching are the only features used by Kripke.
The algorithm performs agglomerative clustering on
document-level entities. A cascade
of fusion steps is performed, where conditions for name and context matching
are slowly relaxed from strict to more generous constraints.
After performing document-level analysis and cross-document coreference,
Kelvin take steps to remove spurious assertions (using hand-written rules and blacklists
to exclude unlikely facts), eliminates extraneous facts (e.g., a person can
only have one city of birth and a reasonable number of children) and facts with
insufficient support, and then performs logical inference to augment the
resulting knowledge base.  The logical inference uses procedurally implemented
forward chaining rules.

\paragraph{Entity Disambiguation}
We have are using two systems for performing entity disambiguation, one designed
to work well when entities are known and another
that performs a more careful analysis to bootstrap a set of
lesser known entities.
\slinky\ \cite{Benton:2014qe} is a streaming entity linker designed to support
both high throughput and low latency linking.  Most work on entity
linking has treated KBP as a batch task; systems typically don't
optimize for speed, meaning batch runs can take hours.
\slinky\ is optimized for IKBP and can quickly produce a
top-$K$ list of entity labels for each mention.

When a set of entities is not known, \parma\ \cite{Wolfe2013}, which
works with topically related pairs documents, constructs an alignment
between the entities and events mentioned.  \parma\ can bootstrap a small KB by
linking new mentions to canonical mentions in an existing document (i.e., a
light-weight KB) or decide to create a new entity if the mentions do not
appear to be coreferent.  Because \parma\ works at the level of pairs of
documents, it can afford to use complex discourse features and joint inference,
which are not feasible at large scale; this makes it ideal for the first stage of IKBP.

\paragraph{Relation Identification}
A large part of IKBP is identifying basic propositions stated in documents.
One version of this task is Semantic Role Labeling (SRL) \cite{Gildea2002}:
identifying and labeling the types of semantic \emph{arguments} tied to
semantic \emph{predicates}.  As a potential user might wish to analyze language
content in low resource languages or domains, we are exploring models for \emph{low-resource} SRL, 
such as that of \newcite{Gormley2014}.
ACE \cite{ACE2004} is another view that comprises entities, values, time
expressions, relations, and events.  The set of relations and events that ACE
annotated are closed-class but high-value, which makes them a nice compromise
between complexity (both annotation and learning) and utility to the user. We
have ongoing work towards building an interactive ACE system.

\section{Conclusion}
IKBP is a new variant of KBP tailored for long-term,
focused information  analysis.  It captures knowledge from a
user's workflow while offering the ability to add rich structure to their
summary reports.
 Much previous work can
be extended to support IKBP systems.
Many aspects of IKBP remain unexplored: 
the annotation UI, which can have a large
effect on an IKBP system's success;
IKBP can benefit from
 more speculative inferences that rely on 
 a human in the loop, as users may write domain-specific
inference rules;
connecting IKBP together across many users, and merging pocket KBs is an open
problem, and merging two clean pocket KBs could produce a more reliable result than
building a global KB.

\bibliographystyle{naaclhlt2015}

\bibliography{sources_short}

\end{document}